\documentclass[final]{UD-Class}

\usepackage[sf,SF]{subfigure}
\usepackage{graphicx}
\usepackage{times}%
\usepackage{lastpage}
\usepackage[sort]{natbib}

\usepackage{url}
\usepackage{amssymb}

\usepackage{xspace}
\usepackage{xcolor}
\usepackage{setspace}

\usepackage{amsmath}
\usepackage{amssymb}
\usepackage[thmmarks]{ntheorem}

\usepackage[bookmarksopen=true,bookmarksnumbered=true,pdftitle={DesignSpace},pdfauthor={Mehul Bhatt},linkbordercolor=blue,urlcolor=red]{hyperref}

\firstpage{1}
\lastpage{\pageref{LastPage}}
\volume{}
\pubyear{2013}
\newcommand{\myurl}[1]{{\upshape\footnotesize\texttt{\url{#1}}}}

\begin{document}
\begin{frontmatter}                           
%


\pretitle{}

\vspace{0.3in}

\title{Between Sense and Sensibility}




\subtitle{Declarative narrativisation of mental models as a basis and benchmark for visuo-spatial cognition and\\computation focussed collaborative cognitive systems}


\runningtitle{Between Sensing and Sensibility}

\medskip

\author{\fnms{\bf Mehul} \snm{Bhatt}} %

\runningauthor{M. Bhatt}

\address{Cognitive Systems, and\\Spatial Cognition Research Center\\
University of Bremen, Germany\\
{\upshape\footnotesize\texttt{\color{black!50!blue} bhatt@informatik.uni-bremen.de}}\\
{\upshape\footnotesize\texttt{\color{black!50!blue} www.mehulbhatt.org}}\\
}

%



\medskip
\medskip
\medskip
\medskip
\medskip
\medskip
\medskip
\medskip
\medskip
\medskip
\medskip
\medskip
\medskip

{\textbf{subject keywords}
\it\

computer science; cognitive science; artificial intelligence; cognitive systems; human-computer interaction
}

\medskip

{\textbf{general keywords}
\it\

computational models of narrative; spatial cognition and computation; commonsense reasoning; spatial and temporal reasoning; action and change; cognitive robotics; design, architecture, urban planning; geographic information systems; spatial assistance systems

}

\end{frontmatter}

$~$
\newpage


$~$

\begin{center}
\huge{\textsc{Position Statement}}
\end{center}

\medskip

What lies between `\emph{sensing}' and  `\emph{sensibility}'? In other words, what kind of cognitive processes mediate sensing capability, and the formation of sensible impressions ---e.g., abstractions, analogies, hypotheses and theory formation, beliefs and their revision, argument formation--- in domain-specific problem solving, or in regular activities of everyday living, working and simply going around in the environment? How can knowledge and reasoning about such capabilities, as exhibited by humans in particular problem contexts, be used as a model and benchmark for the development of collaborative cognitive (interaction) systems concerned with human assistance, assurance, and empowerment?

\smallskip

We pose these questions in the context of a range of assistive technologies concerned with \emph{visuo-spatial perception and cognition} tasks encompassing aspects such as commonsense, creativity, and the application of specialist domain knowledge and problem-solving thought processes \citep{Bhatt-Schultz-Freksa:2013}. Assistive technologies being considered include: (a) human activity interpretation from sensor data; (b) high-level cognitive control for human-robot collaboration; (c) people-centred, function-driven, creative design in domains such as architecture \& digital media creation, and (d) high-level qualitative analyses in spatio-temporal data-intensive geographic information systems

%
%
%
%
%

\medskip

\textbf{A.\quad Compuational Narrativisation as a Benchmark}\quad 

Narrativisation processes pertaining to space, actions, and change are ubiquitous \citep{Tversky-narrative-2004,Bhatt:RSAC:2012}. Humans, robots, and systems involving action and mutual (computer-human) interaction are embedded in \emph{space}. Space, spatial configurations, and in effect, their perceptually grounded \emph{mental models} \citep{Johnson-Laird:1986:Mentalmodels}, declarative abstractions, or ad-hoc system-level representations undergo \emph{change} --- the notion of time, or change-based temporal progression, and therefore \emph{spatio-temporal dynamics} and \emph{spatio-temporal narrativisation} (processes) arise naturally.

\smallskip

The significance of \emph{narratives} in everyday discourse, interpretation, interaction, belief formation, and decision-making has been acknowledged and studied in a range of scientific, humanistic, and artistic disciplines. Narrativisation of everyday perceptions by humans, and the significance of narratives, e.g., in communication and interaction, has been investigated under several frameworks, and through several interdisciplinary initiatives involving the arts, humanities, and social sciences, e.g., the narrative paradigm \citep{narrative-paradigm}, narrative analysis \citep{narrat-analysis}, narratology \citep{narratolog-prince-1982,narratology-marie-laure,meister-narratology-hdbk-narrato}, discourse analysis and computational narratology \citep{Roland-1975-narrative-structural,CMN-Mani-2012,Mani-comp-narratology,Goguen-course-compu-narratology,FSS102323}. Broadly, the study of narratives has attracted attention from several quarters, most prominently in disciplines such as literature, linguistics, anthropology, semiotics, cultural studies, geography, psychology, cognitive science, logic, and computer science. 

\smallskip

We regard narratives, and high-level processes of narrativisation emanating therefrom, as a general underlying structure serving the crucial function of \emph{perceptual sense-making} --- i.e.,  as a link between problem-specific perceptual sensing and the (computational) formation of sensible impressions concerned with interpretation and analytical tasks. Given the nature of the visuo-spatial cognition tasks being considered, the particular form of the proposed narrative structure is that of cognitively inspired \emph{computational model of narrativisation} involving high-level commonsense reasoning with \emph{space, events, actions, change, and interaction}. In particular, the following capabilities are crucial:

\newpage

\begin{enumerate}
{
	\item Narrativising---based on modalities such as language, diagrams, visual abstractions, analogy, conceptual blending, and hypotheses--- certain aspects of perceived reality, e.g., obtained via 2D or 3D perception sensors, from large-scale complex datasets representing spatio-temporal data etc

	\item Achieving the narrativisation ---or human-like perceptual sense-making--- with a level of declarative abstraction,\footnote{Declarative-ness signifies the existence of models that can be reasoned and queried upon, e.g., within a traditional declarative knowledge representation and reasoning framework such as logic programming, constraint logic programming, description logic based conceptual reasoning, answer-set programming, and specialised commonsense reasoners based on expressive action logics. Details of such underlying enabling methods are not reelvant for this position statement.} analytical accuracy, and semantic \& descriptive quality and expressibility that is comparable with, or even outperforming, expected human performance in specific problem contexts.

}	
\end{enumerate}

Computational narratives not only provide a rich cognitive basis, but they also serve as a benchmark of functional performance in our development of computational cognitive assistance systems. We posit that computational narrativisation pertaining to space, actions, and change provides a useful model of  \emph{visual} and \emph{spatio-temporal thinking} within a wide-range of problem-solving tasks and application areas (see B; B1-B5) where collaborative cognitive systems could serve an assistive and empowering function.

\medskip

\textbf{B.\quad Narrativised Spatio-Temporal Thinking in Analytical and Creative Problem Solving}\quad 

The ubiquity and diversity of visuo-spatial narrativisation is self-evident in a range of cognitive (interaction) systems and technologies that aim to assist and empower people in creative, specialist, and regular everyday living situations. Consider assistive technologies and systems concerning the problem domain and application areas in (B1--B5): 


\medskip

{\small\textbf{B1}}.\quad \emph{Perceptual Narratives and Human Activity Interpretation}\quad

Perceptual narratives pertain to artificial sensor grounded visual, auditory, and haptic observations in the real world. Declarative models of perceptual narratives can be used for assistive tasks in the course of activities in everyday life and work (e.g., human activity interpretation, semantic model generation from video, ambient intelligence) \citep{Bhatt2013-CMN,Bhatt2013-Rotunde}.

\smallskip
	
{\small\textbf{B2}}.\quad \emph{Narrative-based Control for Cognitive Robotics}\quad

To the extent that a perceptual narrative, as in {\small (A1)}, can help to make sense of a situation for an artificial agent (e.g., robot, system), it can also be the basis of future action. For instance, a robotic agent can identify abnormalities in a narrative, which may in turn affect subsequent (re)planning and sensing behaviour, and dialog with a user \citep{CR-2013-Narra-CogRob,Suchan-Bhatt-expcog-2014}
\smallskip
					
{\small\textbf{B3}}.\quad \emph{Narratives of User Experience for Spatial Design \& Architecture}

Narratives of user experience pertain to human-grounded visuo-spatial and locomotive modalities. Consider this: you enter a Museum for the first time; as you go around, guided by the internal structure of the building, you form a narrative of user experience based on your visuo-spatial, locomotive, and affordance-based perceptions in the Museum. Architects concerned with designing this Museum are confronted with anticipating this sort of narrative of user experience,  at a time when all that exists during the initial design conception phase is `\emph{empty space}' \citep{bhattschultz2012,DesignNarrative-KR2014}.

\smallskip

{\small\textbf{B4}}.\quad \emph{Narratives  for Creative Assistance in Digital and New Media Production}

We interpret creative assistance in digital media production as the capability of computational tools to support the creative skills of experts and artists at several stages within the media design and creation process. Consider the domain of film, animation, and comic book pre-production. Here, one may identify several forms of assistance at the production phase, e.g., virtual cinematography, storyboarding, and scene visualisation from scripts and automatic camera control in the animation domain. Professionals who may be assisted, in a film context, include: cinematographers, directors, script and screenplay writers, storyboarding artists \citep{Bhatt:STeDy:10}.

\newpage

{\small\textbf{B5}}.\quad \emph{Geospatial Narratives and their Spatio-Temporal Dynamics}

Geospatial narratives attempt to make sense of massive quantities of micro and macro-level spatio-temporal data pertaining to environmental, socio-economic and demographic processes operating in a geospatial context \citep{Bhatt_Wallgruen_11_Analytical,Bhatt-Wallgruen-2014}. Such narratives pertain to spatio-temporal databases of precise measurements about environmental features, aerial imagery, sensor network databases with real-time information about natural and artificial processes and phenomena etc. Geospatial narratives typically span a temporal horizon encompassing generational change, but these could also pertain to the scale of everyday `\emph{life in the city}', natural environmental processes etc.  \citep{Bhatt-Wallgruen-2014}

\smallskip

\emph{In the backdrop of the problem domains in (B1--B5), we are investigating}:\quad  (a) the conceptual and computational aspects of narrative-based visuo-spatial cognition, (b) declarative model of narrative knowledge, and its relationship with spatio-linguistically grounded behavioural and formal theories, (c) role of specialised knowledge representation and reasoning mechanisms as underlying methods for automating high-level narrativisation processes from the viewpoint of visuo-spatial \emph{cognition and creativity}.

\enlargethispage{-15pt}

\section*{Acknowledgements}
I gratefully acknowledge the funding and support of the German Research Foundation (DFG), {\upshape\footnotesize\texttt{www.dfg.de/}} --- part of the work described in this paper has been conducted as part of the DFG funded SFB/TR 8 Spatial Cognition project [DesignSpace],  {\upshape\footnotesize\url{http://www.design-space.org}}. 

\smallskip

I thank Manfred Eppe, Carl Schultz, Jakob Suchan, and Jan Oliver Wallgr\"{u}n for constructive discussions and collaborations related to different aspects of the challenges being addressed / positioned in this abstract.



\bibsep8pt

\renewcommand{\bibname}{Related Work} 
\renewcommand\refname{Related Work} 

\bibliographystyle{abbrvnat}


\begin{thebibliography}{23}
\providecommand{\natexlab}[1]{#1}
\providecommand{\url}[1]{\texttt{#1}}
\expandafter\ifx\csname urlstyle\endcsname\relax
  \providecommand{\doi}[1]{doi: #1}\else
  \providecommand{\doi}{doi: \begingroup \urlstyle{rm}\Url}\fi

\bibitem[Barthes and Duisit(1975)]{Roland-1975-narrative-structural}
R.~Barthes and L.~Duisit.
\newblock {An Introduction to the Structural Analysis of Narrative}.
\newblock \emph{New Literary History}, 6\penalty0 (2):\penalty0 237--272, 1975.
\newblock ISSN 00286087.
\newblock \doi{10.2307/468419}.
\newblock URL \url{http://dx.doi.org/10.2307/468419}.

\bibitem[Bhatt(2012)]{Bhatt:RSAC:2012}
M.~Bhatt.
\newblock {Reasoning about Space, Actions and Change: A Paradigm for
  Applications of Spatial Reasoning}.
\newblock In \emph{Qualitative Spatial Representation and Reasoning: Trends and
  Future Directions}. IGI Global, USA, 2012.
\newblock ISBN ISBN13: 9781616928681.

\bibitem[Bhatt and Flanagan(2010)]{Bhatt:STeDy:10}
M.~Bhatt and G.~Flanagan.
\newblock Spatio-temporal abduction for scenario and narrative completion.
\newblock In M.~Bhatt, H.~Guesgen, and S.~Hazarika, editors, \emph{Proceedings
  of the International Workshop on Spatio-Temporal Dynamics, co-located with
  ECAI-10}, pages 31--36. ECAI Workshop Proceedings., August 2010.

\bibitem[Bhatt and Wallgruen(2011)]{Bhatt_Wallgruen_11_Analytical}
M.~Bhatt and J.~O. Wallgruen.
\newblock Analytical intelligence for geospatial dynamics.
\newblock In \emph{Proceedings of {COSIT 2011}: Conference on Spatial
  Information Theory}, 2011.

\bibitem[Bhatt and Wallgruen(2014)]{Bhatt-Wallgruen-2014}
M.~Bhatt and J.~O. Wallgruen.
\newblock Geospatial narratives and their spatio-temporal dynamics: Commonsense
  reasoning for high-level analyses in geographic information systems.
\newblock \emph{Special Issue on: Geospatial Monitoring and Modelling of
  Environmental Change, ISPRS International Journal of Geo-Information}, 2014.
\newblock ISSN 2220-9964.
\newblock URL \url{http://arxiv.org/abs/1307.2541}.
\newblock In Press.

\bibitem[Bhatt et~al.(2012)Bhatt, Schultz, and Huang]{bhattschultz2012}
M.~Bhatt, C.~Schultz, and M.~Huang.
\newblock {The Shape of Empty Space: Human-centred Cognitive Foundations in
  Computing for Spatial Design}.
\newblock In \emph{IEEE Symposium on Visual Languages and Human-Centric
  Computing (VL/HCC)}, pages 33--40, 2012.

\bibitem[Bhatt et~al.(2013{\natexlab{a}})Bhatt, Schultz, and
  Freksa]{Bhatt-Schultz-Freksa:2013}
M.~Bhatt, C.~Schultz, and C.~Freksa.
\newblock {The `Space' in Spatial Assistance Systems: Conception, Formalisation
  and Computation}.
\newblock In T.~Tenbrink, J.~Wiener, and C.~Claramunt, editors,
  \emph{Representing space in cognition: Interrelations of behavior, language,
  and formal models. Series: Explorations in Language and Space}.
  978-0-19-967991-1, Oxford University Press, 2013{\natexlab{a}}.

\bibitem[Bhatt et~al.(2013{\natexlab{b}})Bhatt, Suchan, and
  Freksa]{Bhatt2013-Rotunde}
M.~Bhatt, J.~Suchan, and C.~Freksa.
\newblock {ROTUNDE} -- {A Smart Meeting Cinematography Initiative}.
\newblock In M.~Bhatt, H.~Guesgen, and D.~Cook, editors, \emph{Proceedings of
  the AAAI-2013 Workshop on Space, Time, and Ambient Intelligence (STAMI).},
  Washington, US, 2013{\natexlab{b}}. AAAI Press.

\bibitem[Bhatt et~al.(2013{\natexlab{c}})Bhatt, Suchan, and
  Schultz]{Bhatt2013-CMN}
M.~Bhatt, J.~Suchan, and C.~Schultz.
\newblock {Cognitive Interpretation of Everyday Activities -- Toward Perceptual
  Narrative Based Visuo-Spatial Scene Interpretation}.
\newblock In M.~Finlayson, B.~Fisseni, B.~Loewe, and J.~C. Meister, editors,
  \emph{Computational Models of Narrative (CMN) 2013., a satellite workshop of
  CogSci 2013: The 35th meeting of the Cognitive Science Society.}, Dagstuhl,
  Germany, 2013{\natexlab{c}}. OpenAccess Series in Informatics (OASIcs).

\bibitem[Bhatt et~al.(2014)Bhatt, Schultz, and Thosar]{DesignNarrative-KR2014}
M.~Bhatt, C.~Schultz, and M.~Thosar.
\newblock Computing narratives of cognitive user experience for building design
  analysis: Kr for industry scale computer-aided architecture design.
\newblock In T.~Eiter, C.~Baral, and G.~D. Giacomo, editors, \emph{Principles
  of Knowledge Representation and Reasoning: Proceedings of the 14th
  International Conference, KR 2014}, 2014.
\newblock URL \url{http://kr.org/KR2014/}.
\newblock to appear.

\bibitem[Eppe and Bhatt(2013)]{CR-2013-Narra-CogRob}
M.~Eppe and M.~Bhatt.
\newblock {Narrative based Postdictive Reasoning for Cognitive Robotics}.
\newblock In \emph{COMMONSENSE 2013: 11th International Symposium on Logical
  Formalizations of Commonsense Reasoning}, 2013.

\bibitem[Fisher(1987)]{narrative-paradigm}
W.~R. Fisher.
\newblock \emph{Human communication as narration: Toward a philosophy of
  reason, value, and action}.
\newblock University of South Carolina Press, Columbia, SC, 1987.

\bibitem[Goguen(2004)]{Goguen-course-compu-narratology}
J.~Goguen.
\newblock {CSE 87C Winter 2004 Freshman Seminar on Computational Narratology.
  }.
\newblock \emph{New Literary History}, 2004.
\newblock URL \url{http://cseweb.ucsd.edu/~goguen/courses/87w04/1.html}.

\bibitem[Herman et~al.(2005)Herman, Jahn, and Ryan]{narratology-marie-laure}
D.~Herman, M.~Jahn, and M.-L. Ryan.
\newblock \emph{Routledge Encyclopedia Of Narrative Theory}.
\newblock Routledge, Feb. 2005.
\newblock ISBN 0415282594.

\bibitem[Johnson-Laird(1983)]{Johnson-Laird:1986:Mentalmodels}
P.~N. Johnson-Laird.
\newblock \emph{Mental models: towards a cognitive science of language,
  inference, and consciousness}.
\newblock Harvard University Press, Cambridge, MA, USA, 1983.
\newblock ISBN 0-674-56882-6.

\bibitem[Lakoff and Narayanan(2010)]{FSS102323}
G.~Lakoff and S.~Narayanan.
\newblock Toward a computational model of narrative, 2010.
\newblock URL
  \url{http://www.aaai.org/ocs/index.php/FSS/FSS10/paper/view/2323}.

\bibitem[Mani(2012)]{CMN-Mani-2012}
I.~Mani.
\newblock Computational modeling of narrative.
\newblock \emph{Synthesis Lectures on Human Language Technologies}, 5\penalty0
  (3):\penalty0 1--142, 2012.

\bibitem[Mani(2013)]{Mani-comp-narratology}
I.~Mani.
\newblock {Computational Narratology}.
\newblock In P.~H\"{u}hn, J.~C. Meister, J.~Pier, and W.~Schmid, editors,
  \emph{{The Living Handbook of Narratology}}. Hamburg University Press, 2013.

\bibitem[Meister(2011)]{meister-narratology-hdbk-narrato}
J.~C. Meister.
\newblock {Narratology}.
\newblock In P.~H\"{u}hn, J.~C. Meister, J.~Pier, and W.~Schmid, editors,
  \emph{{The Living Handbook of Narratology}}. Hamburg University Press, 2011.

\bibitem[Prince(1982)]{narratolog-prince-1982}
G.~Prince.
\newblock \emph{{Narratology: The Form and Function of Narrative}}.
\newblock Mouton, 1982.

\bibitem[Riessman(1993)]{narrat-analysis}
C.~K. Riessman.
\newblock \emph{{Narrative Analysis}}.
\newblock Newbury Park: Sage Publications, 1993.

\bibitem[Suchan and Bhatt(2014)]{Suchan-Bhatt-expcog-2014}
J.~Suchan and M.~Bhatt.
\newblock {The ExpCog Framework: High-Level Spatial Control and Planning for
  Cognitive Robotics}.
\newblock In \emph{Bridges between the Methodological and Practical Work of the
  Robotics and Cognitive Systems Communities - From Sensors to Concepts}.
  Intelligent Systems Reference Library, Springer, 2014.
\newblock (in press).

\bibitem[Tversky(2004)]{Tversky-narrative-2004}
B.~Tversky.
\newblock {Narratives of Space, Time, and Life}.
\newblock \emph{Mind \& Language}, 19\penalty0 (4):\penalty0 380--392, 2004.
\newblock \doi{10.1111/j.0268-1064.2004.00264.x}.
\newblock URL \url{http://dx.doi.org/10.1111/j.0268-1064.2004.00264.x}.

\end{thebibliography}


\end{document}